\documentclass{article}
\PassOptionsToPackage{square, numbers}{natbib}
\usepackage[preprint]{neurips_2026}

\makeatletter
\renewcommand{\@noticestring}{}
\renewcommand{\@toptitlebar}{\vskip 0.1in}
\renewcommand{\@bottomtitlebar}{\vskip 0.15in}
\makeatother

\usepackage[utf8]{inputenc} 
\usepackage[T1]{fontenc}   
\usepackage{hyperref}     
\usepackage{url}           
\usepackage{booktabs}    
\usepackage{amsfonts}     
\usepackage{nicefrac}      
\usepackage{microtype}
\usepackage{xcolor}         
\usepackage{amsmath}
\usepackage{mathtools}
\usepackage{amsthm}
\usepackage{times}
\usepackage{latexsym}
\usepackage{bm} 
\usepackage{graphicx}
\usepackage{amssymb}
\usepackage{amstext}
\usepackage{multirow,multicol}
\usepackage{threeparttable}
\usepackage{makecell}
\usepackage{bbding}
\usepackage{arydshln} 
\usepackage{stfloats}
\usepackage{float}

\usepackage{tikz}
\usetikzlibrary{calc}

\definecolor{msred}{RGB}{242,80,34}
\definecolor{msgreen}{RGB}{127,186,0}
\definecolor{mslogoblue}{RGB}{0,164,239}
\definecolor{msyellow}{RGB}{255,185,0}
\definecolor{msdark}{RGB}{50,50,50}             % near-black for wordmark
\definecolor{mspanelbg}{RGB}{244,244,248}       % refined pearl-gray, subtle cool tone

\newcommand{\microsoftlogo}{%
  \begin{tikzpicture}[baseline=2pt]
    \def\s{0.46cm}\def\g{0.055cm}%
    \fill[msred,rounded corners=0.5pt]      (0,\s+\g) rectangle (\s,2*\s+\g);
    \fill[msgreen,rounded corners=0.5pt]    (\s+\g,\s+\g) rectangle (2*\s+\g,2*\s+\g);
    \fill[mslogoblue,rounded corners=0.5pt] (0,0) rectangle (\s,\s);
    \fill[msyellow,rounded corners=0.5pt]   (\s+\g,0) rectangle (2*\s+\g,\s);
  \end{tikzpicture}%
}

\newcommand{\microsoftlogosmall}{%
  \begin{tikzpicture}[baseline=1pt]
    \def\s{0.30cm}\def\g{0.04cm}%
    \fill[msred,rounded corners=0.4pt]      (0,\s+\g) rectangle (\s,2*\s+\g);
    \fill[msgreen,rounded corners=0.4pt]    (\s+\g,\s+\g) rectangle (2*\s+\g,2*\s+\g);
    \fill[mslogoblue,rounded corners=0.4pt] (0,0) rectangle (\s,\s);
    \fill[msyellow,rounded corners=0.4pt]   (\s+\g,0) rectangle (2*\s+\g,\s);
  \end{tikzpicture}%
}

\newcommand{\microsoftheader}{%
  \begin{tikzpicture}[remember picture, overlay]
    \coordinate (panelNW)
      at ([xshift=0.72in, yshift=-0.28in]current page.north west);
    \coordinate (panelNE)
      at ([xshift=-0.72in, yshift=-0.28in]current page.north east);
    \coordinate (panelSE)
      at ([xshift=-0.72in, yshift=-6.0in]current page.north east);
    \coordinate (stripeL)
      at ([xshift=0.72in, yshift=-0.98in]current page.north west);
    \coordinate (stripeR)
      at ([xshift=-0.72in, yshift=-0.98in]current page.north east);
    \fill[msred]
      (stripeL) rectangle ($ (stripeL)!0.25!(stripeR) + (0,-2.6pt) $);
    \fill[msgreen]
      ($ (stripeL)!0.25!(stripeR) $) rectangle ($ (stripeL)!0.50!(stripeR) + (0,-2.6pt) $);
    \fill[mslogoblue]
      ($ (stripeL)!0.50!(stripeR) $) rectangle ($ (stripeL)!0.75!(stripeR) + (0,-2.6pt) $);
    \fill[msyellow]
      ($ (stripeL)!0.75!(stripeR) $) rectangle ($ (stripeR) + (0,-2.6pt) $);
    \node[anchor=north west, inner sep=0pt]
      at ([xshift=0.98in, yshift=-0.42in]current page.north west) {%
        \microsoftlogo
        \hspace{8pt}%
        {\fontsize{17}{20}\selectfont\sffamily\color{msdark}Microsoft}%
      };
    \node[anchor=south east, inner sep=0pt]
      at ([xshift=-1.22in, yshift=-6.14in]current page.north east) {%
        \microsoftlogosmall
        \hspace{6pt}%
        {\fontsize{11.5}{13}\selectfont\sffamily\color{msdark}Microsoft}%
      };
  \end{tikzpicture}%
}

\title{Speech LLMs are Contextual Reasoning Transcribers}

\author{
  Keqi Deng, Ruchao Fan, Bo Ren, Yiming Wang, Jinyu Li\\
  Microsoft Core AI, USA \\
  \texttt{keqideng@microsoft.com, jinyli@microsoft.com} \\
}

\begin{document}

\microsoftheader
\maketitle

\begin{abstract}
Despite extensions to speech inputs, effectively leveraging the rich knowledge and contextual understanding of large language models (LLMs) in automatic speech recognition (ASR) remains non-trivial, as the task primarily involves direct speech-to-text mapping. To address this, this paper proposes chain-of-thought ASR (CoT-ASR), which constructs a reasoning chain that enables LLMs to first analyze the input speech and generate contextual analysis, thereby fully exploiting their generative capabilities. With this contextual reasoning, CoT-ASR then performs more informed speech recognition and completes both reasoning and transcription in a single pass. Moreover, CoT-ASR naturally supports user-guided transcription: while designed to self-generate reasoning, it can also seamlessly incorporate user-provided context to guide transcription, further extending ASR functionality. To reduce the modality gap, this paper introduces a CTC-guided Modality Adapter, which uses CTC non-blank token probabilities to weight LLM embeddings, efficiently aligning speech encoder outputs with the LLM’s textual latent space. Experiments show that, compared to standard LLM-based ASR, CoT-ASR achieves a relative reduction of 8.7\% in word error rate (WER) and 16.9\% in entity error rate (EER).
\end{abstract}

\section{Introduction}
With the success of text-based large language models (LLMs) and their remarkable performance across a wide range of tasks \cite{brown2020language,touvron2023llama}, they have been extensively extended to directly handle spoken language processing tasks \cite{chu2023qwen, tang2024salmonn, deng2024wav2prompt, abouelenin2025phi}, including automatic speech recognition (ASR) \cite{xu2025fireredasr,bai2024seed,ma2025speech}. Since text-based LLMs have been extensively pre-trained on massive-scale text data that far exceeds what is available for ASR, and since speech and text are semantically connected, integrating LLMs into ASR systems is a promising solution to effectively leverage their rich knowledge and strong contextual understanding capabilities. Existing studies \cite{xu2025fireredasr,bai2024seed,song2025index} have demonstrated the benefits of incorporating LLMs into ASR systems, particularly in resolving semantic ambiguities and generating more coherent transcriptions. ASR is now undergoing a transition from conventional sequence-to-sequence models \cite{li2022recent}, such as attention-based encoder–decoder (AED) models, toward LLM-based architectures \cite{song2025index}.

On the other hand, it is still worth exploring how to fully leverage the capabilities of LLMs in ASR tasks. 
The current paradigm typically prepends the speech encoder output before the text sequence as a prompt \cite{chu2023qwen,ma2025speech, abouelenin2025phi}, allowing the LLM’s self-attention to attend to the speech content when predicting the next token. This merges the functions of self-attention and cross-attention mechanisms used in AED models. 
%The training target of LLM-based ASR systems is still solely the transcription text, which remains the same as conventional AED-based ASR models. 
LLM-based ASR systems are still trained solely on transcription text, as in conventional AED-based ASR.
Moreover, in ASR, speech and transcription largely convey the same information under a noisy source–channel formulation \cite{jelinek1998statistical}, making ASR a content-preserving mapping rather than a true source-to-target semantic transformation. As a result, the LLM in ASR is constrained to reproducing or refining the input, making it less straightforward to translate its strong generative and reasoning capabilities into ASR gains \cite{fathullah2024prompting,deng2025transducer} compared to tasks like speech translation or spoken question answering.
% Moreover, in ASR, speech and transcription largely convey the same information under a noisy source–channel formulation \cite{jelinek1998statistical}, making both source-side representations rather than a true source–target semantic transformation. As a result, the LLM in ASR mainly performs repetition or refinement of the input information during recognition. This makes it less straightforward to translate the strong text-processing capabilities of LLMs, such as those demonstrated in machine translation or text-based question answering, into tangible gains for ASR \cite{fathullah2024prompting,deng2025transducer}, compared to tasks like speech translation or spoken question answering. 
%While many studies focus on optimizing speech encoder training to improve LLM-based ASR performance \cite{bai2024seed,mu2025efficient}, the potential for LLMs to move ASR toward human-level intelligence remains significant. Overcoming the limitations of the current paradigm to more fully exploit LLM capabilities remains a key challenge.
While many studies focus on optimizing speech encoder training to improve LLM-based ASR performance \cite{bai2024seed,mu2025efficient}, how to fully exploit LLM capabilities beyond the current paradigm remains an open challenge.

To this end, this paper proposes CoT-ASR, which is the first work to introduce chain-of-thought (CoT) reasoning into ASR. By constructing a reasoning chain, CoT-ASR enables the speech LLM to transcend the mere generation of verbatim transcriptions. Instead, it leverages the LLM’s extensive internal knowledge and strong contextual understanding to perform analytical reasoning prior to transcription, ultimately yielding higher-quality transcription results. 
This represents a new paradigm for LLM-based ASR, as such generative reasoning capabilities, which is inherent in large-scale pre-trained LLMs, are typically unattainable for conventional sequence-to-sequence ASR models. 
Specifically, upon receiving speech input, CoT-ASR first generates a contextual analysis of the speech content, fully exploiting the knowledge and generative capacity of LLM. Once a comprehensive understanding is established, the CoT-ASR continues to generate and performs transcription.
Importantly, the explicit construction of the chain of thought allows CoT-ASR to perform reasoning and transcription in a one-pass manner, preserving the general design principles and simple generation pipeline of LLMs. Furthermore, while designed for self reasoning, CoT-ASR naturally supports user-guided transcription: users can directly provide context to guide the transcription process and bypass self-generated reasoning.
Leveraging the instruction-following and in-context learning capabilities, CoT-ASR can incorporate such user-provided context to produce higher-quality transcription results. Unless otherwise specified, CoT-ASR operates in its default self-reasoning mode, relying solely on the input speech without using external user-provided context.

In addition, to account for the modality gap between speech and text for the speech LLM, this paper designs a connectionist temporal classification (CTC)-guided Modality Adapter. 
Built upon the speech encoder outputs, the adapter computes CTC \cite{graves2006connectionist} probabilities and explicitly distinguishes between blank and non-blank tokens. The non-blank CTC probabilities at each frame are then used to weight and aggregate LLM embedding vectors, enabling effective modality adaptation. This distinct treatment of blank and non-blank probabilities ensures that information from every encoder frame is effectively utilized, which was found crucial for maintaining model accuracy during preliminary experiments.
Experimental results demonstrate that CoT-ASR substantially improves transcription quality over standard LLM-based ASR, with particularly strong gains in reducing entity error rate (EER), which is critical for preserving key information and improving user experience.

The main contributions of this paper can be summarized in four parts:
\begin{itemize}
    \item CoT-ASR is proposed, which is, to the best of our knowledge, the first reasoning-based ASR model that performs reasoning and transcription in a one-pass manner following chain-of-thought.
    \item Beyond self-reasoned contextual analysis, CoT-ASR naturally supports user-guided transcription, extending the functionality and versatility of LLM-based ASR.
    \item A CTC-guided Modality Adapter is designed for effective speech-to-text modality adaptation by virtue of CTC-based probabilistic alignment.
    \item CoT-ASR delivers superior transcription and naturally enhances entity recognition via contextual reasoning, while preserving general LLM design and standard generation pipeline.
    %CoT-ASR delivers superior transcription and naturally enhances entity recognition via its inherent contextual reasoning, keeping a standard LLM generation pipeline. %preserving the model’s generality.
\end{itemize}

The rest of the paper is organized as follows. Section~\ref{sec:related} reviews related work and discusses differences to CoT-ASR. Section~\ref{sec:method} describes CoT-ASR in detail. Section~\ref{setup} and~\ref{sec:results} detail the experimental setup and results, followed by conclusion in Section~\ref{conclusion}.

\section{Related Work}
\label{sec:related}
\subsection{Speech LLMs}

%paragraph 1: text LLM -> adapter: linear, q-former, ctc-compressor, cif, ... for speech llm, HAT
%paragraph 2: speech llm: general speec llm :qwen-audio -> llm-based asr
The landscape of text-based LLMs has undergone a profound transformation, moving beyond simple parameter scaling toward sophisticated reasoning and multi-modal integration. Early milestones such as GPT-3 \citep{brown2020language} and PaLM \citep{chowdhery2023palm} set the foundation for this revolution, which has culminated in frontier models like GPT-4o \citep{hurst2024gpt} and Gemini 1.5 \citep{team2024gemini}, demonstrating exceptional zero-shot capabilities across complex reasoning tasks. 
Concurrently, the focus has shifted toward democratizing high-performance AI through high-quality ``small'' language models (SLMs). Models such as Llama 3 \citep{grattafiori2024llama} and the Phi-4 series \citep{abouelenin2025phi} greatly advance the performance–efficiency trade-off, achieving results comparable to much larger models while maintaining high computational efficiency.
%Models such as Llama 3 \citep{grattafiori2024llama} and the Phi-4 series \citep{abouelenin2025phi} have established a new Pareto frontier, achieving performance comparable to significantly larger models while maintaining high computational efficiency. 
%These advancements in LLMs provide a robust semantic backbone for reasoning-augmented speech recognition frameworks.

The success of text-based LLMs has been extensively extended to speech tasks. Generally, Speech LLMs can be categorized into two paradigms \cite{DBLP:conf/acl/CuiYJMZWGK25}: 1) expanding the LLM vocabulary with discrete speech tokens \cite{zhang-etal-2023-speechgpt, borsos2023audiolm, wang2023neural}, and 2) utilizing continuous speech representations as soft prompts for the LLM \cite{chu2023qwen,ma2025speech, abouelenin2025phi}. This work adopts the latter approach, as it has demonstrated superior performance in ASR tasks. To bridge the inherent modality gap between speech and text, various research efforts have focused on adapting speech encoder outputs into the LLM latent space using modules such as Q-former \cite{tang2024salmonn}, CIF \cite{deng2024wav2prompt,simuls2s-llm-deng-2025}, or CTC-compressors \cite{wu2023decoder,DBLP:conf/eacl/GaidoCNT21, fan2025alignformer}. Among these, simple linear projections \cite{abouelenin2025phi,ma2025speech,mu2025efficient} remain the widely used adapters, as preserving the uncompressed temporal information of the encoder output often yields good performance on downstream tasks. 
While our proposed CTC-guided Modality Adapter also leverages the CTC mechanism, which is related to the CTC-compressor, it does not reduce the temporal length of the encoder outputs. Furthermore, it explicitly utilizes pre-trained LLM text embeddings to facilitate the modality transformation. In addition, the proposed adapter distinguishes between blank and non-blank probabilities in a manner related to HAT \cite{Variani2020HybridAT}; however, HAT is designed for neural transducer models and has different motivation. 
With the adapter facilitating modality adaptation, LLM-based ASR models have been widely established, achieving strong performance across various benchmarks.

\subsection{CoT Reasoning}
%introduce cot -> reasoning LLM

CoT \citep{wei2022chain} enables LLMs to decompose complex problems into explicit intermediate reasoning steps. By modeling reasoning processes, CoT substantially improves performance on tasks involving multi-step inference, compositional reasoning, and contextual knowledge integration.
Early studies mainly explored CoT through specialized prompting techniques \citep{wei2022chain} or inference-time search algorithms \citep{hao2023reasoning, snell2025scaling}. 
Recent studies have shifted toward internalizing reasoning abilities within model parameters via supervised fine-tuning on high-quality, reasoning-augmented data, as exemplified by models such as OpenAI o1.

% Chain-of-Thought (CoT) reasoning~\citep{wei2022chain} enables LLMs to decompose complex problems into explicit intermediate steps, improving multi-step inference, compositional reasoning, and contextual understanding. Early work focused on prompting~\citep{wei2022chain} or inference-time search methods like Monte Carlo Tree Search~\citep{hao2023reasoning, snell2025scaling}. More recently, reasoning capabilities have been internalized through supervised fine-tuning on high-quality, reasoning-augmented data, as exemplified by models such as OpenAI o1.

More recently, CoT reasoning has been extended beyond purely textual tasks to multi-modal and speech-related settings. %LLaVA-CoT \cite{Xu_2025_ICCV} optimizes inference-time sampling and search strategies to refine reasoning paths. Subsequent studies, such as Virgo \cite{du2025virgo}, investigated the transfer of reasoning capabilities from text to vision through structured data organization. The scope has further expanded to video and audio-visual understanding: 
MAmmoTH-VL~\cite{guo-etal-2025-mammoth} introduces large-scale instruction-tuning data to enhance cross-modal question answering, and video-SALMONN-o1~\cite{sun2025videosalmonno1} focuses on general video understanding scenarios.
%In the speech domain, CoT has been applied to speech translation by decomposing it into ASR followed by translation~\citep{du2024cot}, and to speech emotion recognition to enhance interpretability and robustness~\citep{zhao2025steering}. 
%However, in ASR tasks, effectively leveraging CoT reasoning to fully harness LLM capabilities for improved recognition remains a challenge.
In the speech domain, CoT has been explored in several directions. It has been applied to speech translation by decomposing it into ASR followed by translation~\citep{du2024cot}, and to speech emotion recognition to improve interpretability and robustness~\citep{zhao2025steering}. Some recent works further investigate reasoning-style modeling in speech-related settings. 
For example, CoT-based ASR error correction \cite{yang2025chain} adopts a two-pass pipeline, where an initial ASR hypothesis is first generated and then refined, which differs from the one-pass reasoning paradigm typically of LLMs and our CoT-ASR. 
Other works, such as Audio-Reasoner \citep{zhifei-etal-2025-audio} and Step-Audio-R1 \cite{tian2025step}, are not designed or evaluated for ASR tasks and do not perform reasoning-based ASR.
Therefore, effectively leveraging CoT reasoning to improve ASR remains challenging, as speech-to-text is largely content-repetitive with limited semantic shift in ASR, making it difficult to design effective reasoning chains.
% However, for ASR tasks, how to effectively utilize CoT reasoning to fully exploit the capabilities of LLMs for improving ASR performance remains a challenge.

%\newpage
\begin{figure*}[th!]
    \centering
    %\vspace{-0.2cm}
    \includegraphics[width=125mm]{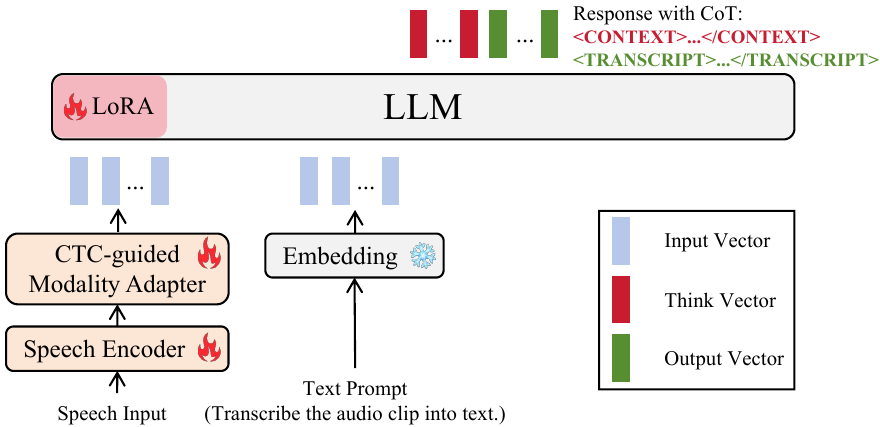}
    %\vspace{-0.06cm}
    \caption{Illustration of CoT-ASR. The text prompt is a fixed template for the ASR task. The output sequence comprises contextual analysis and transcription, marked by the \texttt{<CONTEXT>} and \texttt{<TRANSCRIPT>} tags, respectively. The reasoning-based contextual analysis is highlighted in red, while the subsequent transcription is marked in green. Both are produced sequentially in a single one-pass generation.}
    %\vspace{-0.15cm}
    \label{CoT-ASR}
\end{figure*}
\section{CoT-ASR}
\label{sec:method}

This paper proposes CoT-ASR, a new ASR paradigm for LLM-based ASR to leverage the strong textual capabilities of LLMs. As shown in Fig.~\ref{CoT-ASR}, given speech input together with a text prompt, CoT-ASR differs from standard LLM-based ASR systems that directly generate transcription text. Instead of immediately producing the transcription, CoT-ASR first performs contextual analysis and reasoning over the input speech. This design allows the LLM’s generation is no longer confined to verbatim transcription; rather, it fully exploit its rich knowledge and powerful generative ability to infer the underlying context. This intermediate output represents the model’s reasoning process, highlighted in red in Fig.~\ref{CoT-ASR}. Following a comprehensive understanding of the context, CoT-ASR then proceeds to generate the final transcription, marked in green in Fig.~\ref{CoT-ASR}. Notably, both reasoning and transcription are produced in a single one-pass generation.

\subsection{CoT-ASR Architecture}
The architecture of CoT-ASR is illustrated in Fig.~\ref{CoT-ASR}, comprising three main components: a speech encoder, a CTC-guided Modality Adapter, and an LLM (including its embedding layer). 
Denote the speech encoder output as $\mathbf{E} = (\mathbf{e}_{1}, \dots, \mathbf{e}_{L})$, where $L$ represents the sequence length. The CTC-guided Modality Adapter further transforms $\mathbf{E}$ into a modality-aligned sequence $\mathbf{A} = (\mathbf{a}_{1}, \dots, \mathbf{a}_{L})$ of the same length. 
The text prompt is mapped to an embedding sequence \(\mathbf{I}\) through the LLM embedding layer, which is then concatenated with the speech prompt \(\mathbf{A}\) and fed into the LLM as the generation prompt.

This design strictly adheres to the standard next-token prediction paradigm of LLMs. Given that LLM generation frameworks have been widely developed and optimized by the community (e.g., HuggingFace \cite{wolf-etal-2020-transformers} and vLLM \cite{kwon2023efficient}), maintaining this general generation paradigm is crucial for the compatibility and scalability of CoT-ASR.
Suppose the output sequence (comprising both reasoning and transcription) as $\mathbf{y} = (y_{1}, \dots, y_{N})$, the generation process of CoT-ASR can be formulated as follows:
\begin{equation}
P(\mathbf{y} | \mathbf{E}, \mathbf{I}) = \prod_{i=1}^{N} P(y_i | [\mathbf{A}; \mathbf{I}], y_{<i})
\label{eq:generation}
\end{equation}
where $[\cdot ; \cdot]$ denotes the concatenation operation along the temporal dimension, and $y_{<i}$ represents the tokens generated in previous steps.

%----------
\subsection{Training}
\label{sec:train}

% The training of CoT-ASR follows a supervised fine-tuning paradigm, centered on the construction of speech-grounded reasoning chains and joint multi-task optimization.
The training of CoT-ASR involves two key stages: constructing speech-grounded reasoning chains and supervised fine-tuning with reasoning-augmented speech-text data.

% \paragraph{Chain-of-Thought Data Construction} 
The core of the training strategy involves transforming static transcriptions into dynamic reasoning paths, thereby establishing an explicit CoT for ASR. Given an ASR training dataset, an off-the-shelf text-based LLM is employed to generate a contextual analysis for each ground-truth transcription. This analysis functions as a reasoning intermediate, capturing high-level semantic understanding, disambiguation cues, and inferred background knowledge. 
As illustrated in Fig.~\ref{CoT-ASR}, the final target sequence $\mathbf{y}$ is structured to enforce a reason-before-transcribe logic: 
\texttt{<CONTEXT>} $\text{Contextual Analysis}$ \texttt{</CONTEXT>} \texttt{<TRANSCRIPT>} $\text{Verbatim Transcript}$ \texttt{</TRANSCRIPT>}. 
By training on these reasoning-augmented targets, CoT-ASR learns to perform chain-of-thought reasoning over the speech content before generating the transcription, effectively grounding the output in both inferred context and input speech.

% \subsection{Training}

% The training of CoT-ASR involves two key stages: data preparation with reasoning chains and supervised fine-tuning with reasoning-augmented speech-text data.

% Given an ASR training dataset with speech input and ground-truth transcription, an off-the-shelf text-based LLM is used to generate a contextual analysis for each transcription. This analysis captures high-level semantic understanding, disambiguation cues, and relevant background knowledge inferred from the transcript.
% The final target sequence $\mathbf{y}$ is formatted as a structured string: \texttt{<CONTEXT>} $\text{Contextual Analysis}$ \texttt{</CONTEXT>} \texttt{<TRANSCRIPT>} $\text{Verbatim Transcript}$ \texttt{</TRANSCRIPT>}. 
% By training on this augmented target, CoT-ASR learns to reason about the contextual information while remaining grounded in the speech input and guide the transcription process.

At training, the primary objective is the cross-entropy (CE) loss for the auto-regressive generation of the target sequence $\mathbf{y}$. Given the prefix $[\mathbf{A}; \mathbf{I}]$, the CE loss is defined as:
\begin{equation}
\mathcal{L}_{\text{ce}} = - \sum_{i=1}^{N} \log P(y_i | [\mathbf{A}; \mathbf{I}], y_{<i})
\label{eq:ce_loss}
\end{equation}
In addition, the CTC-guided Modality Adapter (see Section~\ref{sec:ctc-adapter}) in CoT-ASR is trained with a CTC objective computed from the speech encoder outputs. Therefore, the final training loss is a weighted combination of the CE loss and the CTC loss:
\begin{equation}
\mathcal{L} = \mathcal{L}_{\text{CE}} + \lambda \mathcal{L}_{\text{CTC}} \label{eq:loss}
\end{equation}
where $\lambda$ is a hyper-parameter.

\begin{figure*}[t!]
    \centering
    %\vspace{-0.2cm}
    \includegraphics[width=138mm]{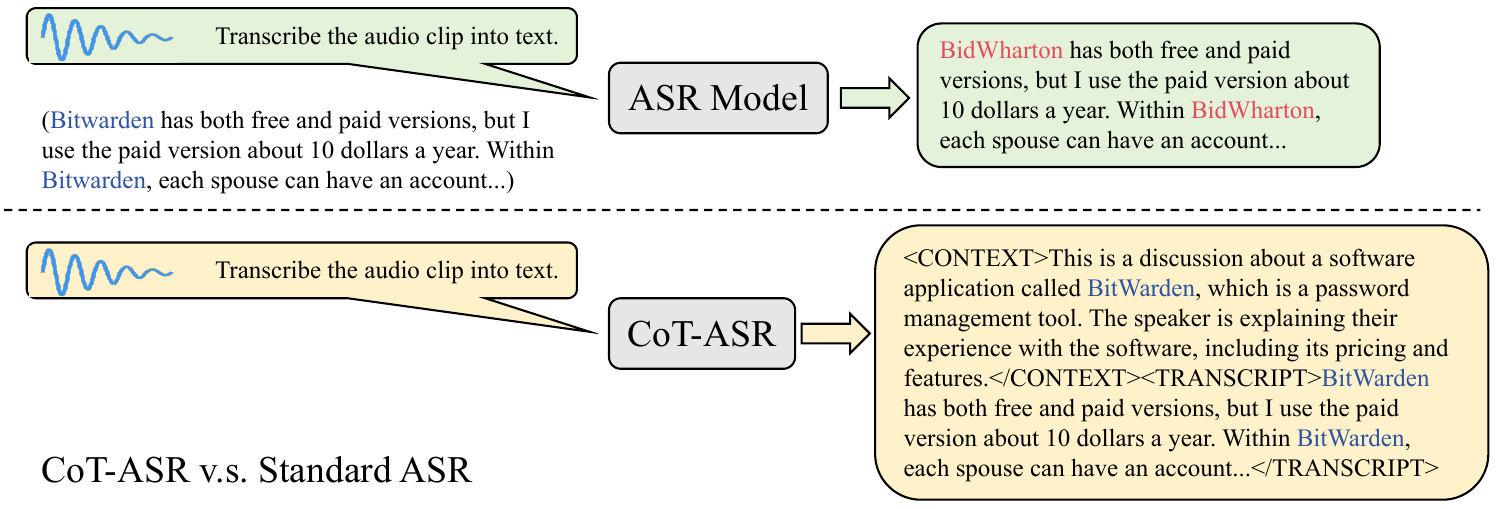}
    %\vspace{-0.06cm}
    \caption{Comparison between CoT-ASR and standard LLM-based ASR. During generation, CoT-ASR first performs a contextual reasoning analysis, which then guides the subsequent transcription. This example illustrates how CoT-ASR leverages the rich knowledge of LLMs to ultimately improve transcription quality.}
    %\vspace{-0.15cm}
    \label{CoT-ASR-egs}
\end{figure*}

\subsection{Inference}
After training, given an input speech signal together with a fixed text prompt instruction, CoT-ASR begins generation and completes both reasoning and transcription within a single pass. Fig.~\ref{CoT-ASR-egs} presents an example of CoT-ASR generation and compares it with standard LLM-based ASR. Through large-scale pretraining, text-based LLMs acquire rich knowledge, such as knowledge about \textit{BitWarden} in the example shown in Fig.~\ref{CoT-ASR-egs}. During the reasoning stage, CoT-ASR effectively leverages this knowledge to analyze, interpret, and disambiguate the speech content, correctly inferring that the context centers around \textit{BitWarden}. The inferred contextual understanding then guides the subsequent transcription, leading to improved transcription quality.

\paragraph{User-guided Transcription}: Beyond self-reasoned inference, CoT-ASR inherently supports user-guided transcription, allowing users to direct the process by providing explicit contextual information. For instance, when a \textit{User Context} is provided, it is encapsulated within the \texttt{<CONTEXT>} \textit{User Context} \texttt{</CONTEXT>} tags and concatenated with the speech prompt $\mathbf{A}$ and the fixed template $\mathbf{I}$ to serve as the unified generation prompt. In this scenario, CoT-ASR bypasses the self reasoning phase and proceeds directly to text transcription. Leveraging its instruction-following and in-context learning capabilities, CoT-ASR effectively incorporates the user-provided context to produce higher-quality transcriptions.
It is worth noting that using user context in this way is only an additional capability of CoT-ASR and not the default setting. By default, CoT-ASR performs context analysis through its own reasoning based solely on the input speech. Unless otherwise specified, all results in this paper rely only on speech input without using external context.

\subsection{CTC-guided Modality Adapter}
\label{sec:ctc-adapter}
\begin{figure}[h]
    \centering
    %\vspace{-0.2cm}
    \includegraphics[width=70mm]{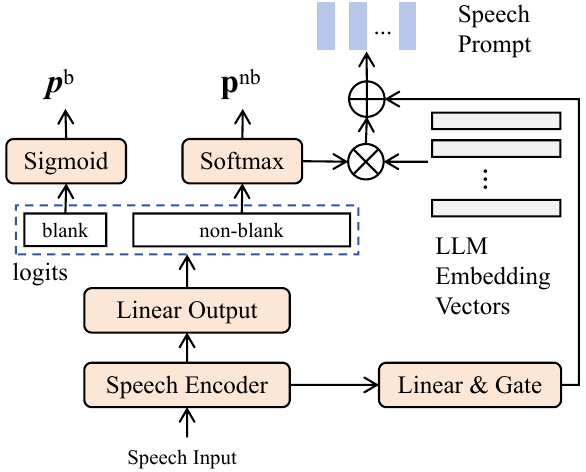}
    %\vspace{-0.25cm}
    \caption{Illustration of the CTC-guided Modality Adapter. The linear output layer projects the encoder outputs to the CTC vocabulary size including the blank token, while the linear layer maps the encoder outputs to the LLM hidden dimension. $\bm{\otimes}$ denotes matrix multiplication and $\bm{\oplus}$ denotes addition. Gate denotes a frame-wise scalar from the linear projection and sigmoid that modulates the residual contribution.}
    %\vspace{-0.2cm}
    \label{fig:ctc-adapt}
\end{figure}

While many LLM-based ASR systems \cite{abouelenin2025phi,ma2025speech,mu2025efficient} employ simple linear projections (optionally with a non-linear activation) as a modality adapter to implicitly align speech encoder outputs with the LLM textual latent space, this paper proposes the CTC-guided Modality Adapter. This component facilitates a more explicit and efficient modality adaptation by leveraging frame-level CTC probability distributions. 

As shown in Fig.~\ref{fig:ctc-adapt}, given the speech encoder output $\mathbf{E} = (\bm{e}_1, \dots, \bm{e}_L)$, a linear output layer first projects each frame representation to logits over a CTC vocabulary that includes the blank token. For each frame, the blank probability is computed using a sigmoid function:
\begin{equation}
p_t^{b} = \text{Sigmoid}(z_t^{b}) \label{eq:b}
\end{equation}
where $z_t^{b}$ denotes the blank logit at frame $t$. For the non-blank tokens, a softmax is applied to obtain a normalized distribution:
\begin{equation}
{\bm{p}}_t^{nb} = \text{Softmax}(\bm{\bm{z}}_t^{nb})
\end{equation}
where ${\bm{p}}_t^{nb} \in \mathbb{R}^{V}$ and $V$ is the vocabulary size (excluding blank token).
To ensure proper normalization when computing CTC loss, the final non-blank probability distribution $\tilde{\bm{p}}_t^{nb}$ is scaled by the non-blank mass:
\begin{equation}
\tilde{\bm{p}}_t^{nb} = (1 - p_t^{b}) \cdot {\bm{p}}_t^{nb} \label{eq:b_hat}
\end{equation}
such that the total probability mass over blank and non-blank tokens sums to one. 
%$\tilde{\bm{p}}_t^{nb}$ is only used for CTC loss computation.

%CTC distributions are known to be spiky \cite{graves2006connectionist}, with most frames dominated by blank predictions. By explicitly separating the computation of blank and non-blank probabilities, the proposed adapter is able to exploit information from every frame, including those dominated by blank predictions, while still producing a meaningful non-blank distribution.

CTC distributions are known to be spiky \cite{graves2006connectionist}, with most frames dominated by blank predictions. By explicitly separating the computation of blank and non-blank probabilities, the proposed adapter is able to exploit information from every frame, including those dominated by blank predictions, while still producing a meaningful non-blank distribution. This design is particularly important, as prior work \cite{fathullah2024prompting} and our large-scale pre-experiments show that compressing the speech prompt, e.g., by discarding blank-dominant frames, often leads to degradation in ASR performance.

To perform modality adaptation, the non-blank probability distribution ${\bm{p}}_t^{nb}$ is used to compute a weighted combination of the LLM token embedding matrix $\mathbf{W}_{\text{emb}} \in \mathbb{R}^{V \times D}$ ($D$ is the hidden dimension):
\begin{equation}
\bm{u}_t = {\bm{p}}_t^{nb} \bm{\otimes} \mathbf{W}_{\text{emb}}
\end{equation}
where $\bm{\otimes}$ denotes matrix multiplication. The resulting vector $\bm{u}_t$ represents the modality-aligned speech representation at frame $t$. 
%A heuristic threshold $\tau$ is applied to $\tilde{\bm{p}}_t^{nb}$, where probabilities below the threshold are set to zero to suppress low-confidence activations.
A heuristic threshold $\tau$ is applied element-wise to ${\bm{p}}_t^{nb}$, such that entries with probability lower than $\tau$ are set to zero to suppress low-confidence activations.

In the extreme case where ${\bm{p}}_t^{nb}$ collapses to a one-hot distribution, the adapted representation $\bm{u}_t$ exactly corresponds to the embedding vector of the selected LLM token, providing an intuitive interpretation of the adapter.

In addition, a gated residual branch is applied to the encoder output $\mathbf{E}$ to perform residual learning.
Specifically, a linear projection maps $\bm{e}_t$ to a $(D+1)$-dimensional vector, where the first $D$ dimensions form the residual representation and the last dimension is passed through a sigmoid function to produce a frame-wise scalar gate $g_t \in (0,1)$.
The final speech prompt is computed as:
\begin{equation}
\bm{a}_t = \bm{u}_t + g_t \cdot \text{Linear}(\bm{e}_t)
\label{eq:residual}
\end{equation}
% In addition, a gated residual branch is applied to the encoder output $\mathbf{E}$ to perform residual learning.
% Specifically, a linear projection followed by a sigmoid activation produces a scalar gate $g_t \in (0, 1)$ for each frame, which modulates the residual contribution from the encoder.
% The final speech prompt is computed as:
% \begin{equation}
% \bm{a}_t = \bm{u}_t + g_t \cdot \text{Linear}(\bm{e}_t),
% \label{eq:residual}
% \end{equation}
% In addition, a separate linear layer is applied directly to the encoder output $\mathbf{E}$ to perform residual learning. The final speech prompt $\mathbf{A}$ is obtained by:
% \begin{equation}
% \bm{a}_t = \bm{u}_t + \text{Linear}(\bm{e}_t)
% \label{eq:residual}
% \end{equation}
The resulting sequence $\mathbf{A} = (\bm{a}_1, \dots, \bm{a}_L)$ is then fed into the LLM as the speech prompt. By fully exploiting CTC probability distributions, the proposed CTC-guided Modality Adapter enables efficient and effective speech-to-text modality adaptation while preserving fine-grained frame-level information.

\section{Experimental Setup}
\label{setup}
\subsection{Dataset}
\label{sec:data}
Experiments were conducted on 38,000 hours anonymized in-house English ASR data. Qwen2.5-14B-Instruct \cite{qwen2} was used to construct reasoning-augmented speech–text data from this dataset. The prompt templates are listed in Appendix~\ref{app:prompt}.
CoT-ASR was evaluated on the English ASR test sets of publicly available benchmarks, including LibriSpeech \cite{7178964} and FLEURS \cite{conneau2023fleurs}, as well as on a series of real-world industry evaluation sets across multiple domains, including consumer goods, nutrition, gaming, wellness, patient history, pharmacy, surgery, and banking.

\subsection{Model Descriptions}
Phi4-mini-instruct \cite{abouelenin2025phi} was employed as the LLM backbone in this paper. Phi4-mini is a compact LM with 3.8B parameters, while achieving performance comparable to larger 7B-scale models such as LLaMA \cite{touvron2023llama}.
The speech encoder was based on a Conformer \cite{gulati20_interspeech} architecture, consisting of three convolutional layers followed by 24 Conformer blocks. Each block employed an attention dimension of 1024, a feed-forward dimension of 1536, and 16 attention heads. 
% The convolutional frontend performs temporal down-sampling with a factor of 8, yielding an effective frame stride of 80 ms at the output of the encoder.
The convolutional frontend down-samples the input by a factor of 8, resulting in an 80 ms encoder output frame stride.

%The convolutional frontend performs temporal down-sampling with a factor of 8, resulting in an effective token rate of 80 ms for the modality adapter.
In addition to CoT-ASR, a Phi-4-MultiModal (Phi4MM)~\cite{abouelenin2025phi} style baseline was constructed in this paper to enable a fair comparison with our built CoT-ASR. Both models used the same speech encoder, LLM backbone, and training strategy.
\paragraph{CoT-ASR:} In addition to the encoder and LLM described above, CoT-ASR employed the proposed CTC-guided Modality Adapter, with threshold $\tau$ set to 0.05. Specifically, the linear layers illustrated in Fig.~\ref{fig:ctc-adapt} consisted of two feed-forward projections separated by a GELU \cite{hendrycks2016gaussian} activation function. $\lambda$ in Eq.~\ref{eq:loss} was set to 0.5. CoT-ASR was trained on the reasoning-augmented ASR data constructed as described in Section~\ref{sec:train}. When evaluating the performance of user-guided transcription of CoT-ASR, Qwen2.5-14B-Instruct \cite{qwen2} was used to extract contextual information as user-provided context.

\paragraph{Phi4MM-style Baseline:} The same encoder and LLM backbone as in CoT-ASR were employed, following the design principles described in \citet{abouelenin2025phi}. As in \citet{abouelenin2025phi}, two feed-forward projections with a GELU activation function was utilized as the adapter. The Phi4MM-style baseline was constructed as a standard LLM-based ASR model and was trained on the ASR datasets as introduced in Section~\ref{sec:data}.

With a batch size of 230k tokens, including both speech encoder output frames and text tokens, both models were trained in two stages. In the first stage, only the adapter was updated for 2,000 steps. In the second stage, the encoder and adapter were updated jointly, and the LLM was fine-tuned via LoRA for 15,000 steps.
%Both models were trained with 10k steps with a batch size of 280 k tokens, including speech encoder output frames and text tokens. 
AdamW was employed as the optimizer with a linear learning rate decay schedule, peaking at 4e-5. The learning rate was warmed up during the first 100 steps. DeepSpeed ZeRO \cite{DBLP:conf/sc/RajbhandariRRH20} stage 1 was used.
Appendix~\ref{sec:efficiency} gives details about inference efficiency.

\subsection{Metrics}
%Word error rate (WER) is used to evaluate the performance of ASR models. However, WER is no longer the sole indicator for evaluating recognition performance \cite{song2025index} as it does not fully capture all aspects of ASR performance, and the recognition of certain entities is particularly important for user experience, entity error rate (EER) is more broadly used in this paper to assess ASR performance. EER is computed in a manner similar to WER, but only counts errors in named entities.

Word Error Rate (WER) remains the conventional metric for evaluating ASR systems; however, it often fails to capture recognition errors that are most critical to real-world user experience. In practical applications, the misrecognition of key entities such as medical terms or personal names typically has a far greater impact on usability than errors involving common function words, a distinction that standard WER is unable to adequately reflect \cite{song2025index}. To address this, this paper primarily adopt the Entity Error Rate (EER) as our evaluation metric. Defined as $1 - \text{Entity Recall}$, EER measures the proportion of reference entities that are not correctly recognized in their entirety. Under this metric, an entity is considered correctly transcribed only if all of its constituent words are error free, thereby enforcing a strict evaluation of the semantic integrity of critical information.
Moreover, EER is particularly suitable for evaluating the effective use of LLMs, as our large-scale pre-experiments indicate that LLM-based ASR shows noticeable improvements in entity recognition compared to traditional ASR systems.

\section{Experimental results}
\label{sec:results}
This section compares the proposed CoT-ASR with the constructed Phi4MM-style baseline. The experiments also demonstrate the performance gains achieved by CoT-ASR when incorporating user-provided context. Except for this section, all other results are obtained using only speech as input, without any user-provided context. In addition, ablation studies are conducted to evaluate the effectiveness of the CTC-guided Modality Adapter. Finally, CoT-ASR is compared experimentally with a range of strong open-source large-scale models.

\subsection{Main ASR Results}
% \begin{table*}[ht!]
% %\vspace{-0.25cm}
%   \caption{WER results on English ASR test sets from public datasets, including LibriSpeech (test-clean and test-other) and FLEURS, along with EER results on a series of in-house English ASR evaluation sets spanning diverse domains. domains. Note that the training data used for the models constructed in this paper do not include these public datasets.}
%   %\vspace{-0.05cm}
%   \label{tab:asr:main}
%   \centering
%   \setlength{\tabcolsep}{2mm}
%   \renewcommand\arraystretch{1.1}
%   \begin{tabular}{l|c| c| c |c|c|c|c|c|c|c|c|c}
%     \Xhline{3\arrayrulewidth}
%     %\multirow{2}{*}{ST Model}&Quality&{Inference}&\multirow{2}{*}{Speedup}&Inference GPU&GPU Memory\\
%     %&(BLEU)&Time (s)&&Memory (MiB)&Reduction\\
%     \multirow{2}{*}{ASR Model} & \multicolumn{3}{c|}{WER} & \multicolumn{8}{c}{EER} \\
%     \cline{2-6}
% &test-clean&test-other&FLEURS&Nutrition&Wellness&Clinical Meetings&Patient History&Pharmacy&Surgery&Banking&Medical\\
%     \hline
%     %NeurST \cite{DBLP:conf/acl/ZhaoWDYL21}&22.8&---&---&---&---\\
%     %Fairseq-ST \cite{DBLP:conf/ijcnlp/WangTMWOP20}&22.7&---&---&---&---\\
%     MHA&23.18&281.3&1.00$\times$&18646&1.00\\
%     MLA&22.97&97.0&2.90$\times$&5065&3.68\\
%     \Xhline{3\arrayrulewidth}
%   \end{tabular}
% %\vspace{-0.08cm}
% \end{table*}

\begin{table}[ht!]
  \caption{WER ($\downarrow$) results on English ASR test sets from public datasets, including LibriSpeech (test-clean and test-other) and FLEURS, along with EER ($\downarrow$) results on a series of in-house English ASR evaluation sets spanning diverse domains. Note that the training data used for the models constructed in this paper do not include these public datasets.}
  \label{tab:asr:main}
  \centering
  \setlength{\tabcolsep}{8pt}
  \renewcommand\arraystretch{1.1}
  \begin{tabular}{c|l|c|c}
    \Xhline{3\arrayrulewidth}
    \multirow{2}{*}{Metric} & \multirow{2}{*}{Test Set} & Phi4MM-style& \multirow{2}{*}{CoT-ASR} \\
    &&Baseline\\
    \Xhline{2\arrayrulewidth}

    \multirow{3}{*}{WER}
      & LibriSpeech test-clean & 2.41 & 2.20 \\
      & LibriSpeech test-other & 5.10 & 4.82 \\
      & FLEURS                & 3.85 &3.57\\

    \hline

    \multirow{8}{*}{EER}
      & Consumer Goods  &15.45&13.46\\
      & Nutrition           & 7.36 & 6.66 \\
      &Gaming&17.42&14.53\\
      & Wellness            & 14.27  & 11.51\\
      % & Clinical Meetings   & 5.63  &  5.21  \\
      & Patient History     & 4.31    & 3.51  \\
      & Pharmacy            & 9.01   & 5.97 \\
      & Surgery             & 7.44   & 6.15  \\
      & Banking             & 12.99   & 11.57  \\
      % & Medical             & 10.65   & 9.47  \\
      \cline{2-4}
      &Average&11.03&9.17\\

    \Xhline{3\arrayrulewidth}
  \end{tabular}
\end{table}

The experimental results are summarized in Table~\ref{tab:asr:main}, which compares CoT-ASR with the Phi4MM-style baseline model, a standard LLM-based ASR model. Table~\ref{tab:asr:main} reports ASR results on public datasets as well as on a series of in-house English ASR evaluation sets from different domains. The results show that although the training data used in this paper do not include the public datasets, making these test sets out-of-domain for the models built in this paper, the two built models still demonstrate robust performance. Moreover, CoT-ASR outperforms the Phi4MM-style baseline model even on these relatively simple public benchmarks, achieving, for example, a relative WER reduction of 8.7\% on LibriSpeech test-clean. 
The primary focus of the experiments in this paper is on the EER results from the in-house ASR evaluation sets, as they not only better reflect real-world ASR user experience but also more effectively assess the ability of LLM-based ASR systems to leverage the knowledge of LLMs. Corresponding WER results on the same in-house test sets are reported in Appendix~\ref{sec:appendix_internal_wer}.
%The primary focus of the experiments in this paper is on the EER results from the in-house ASR evaluation sets, as they better reflect real-world ASR user experience. 
As shown in Table~\ref{tab:asr:main}, CoT-ASR consistently yields great improvements over the Phi4MM-style baseline model across different domains, ultimately achieving a relative reduction of 16.9\% in the average EER. 
These entities represent the knowledge acquired by the LLM during its large-scale pre-training phase, and compared to standard LLM-based ASR systems, CoT-ASR is better able to leverage the capabilities of LLMs through reasoning-based generation.

% \begin{table*}[ht!]
% %\vspace{-0.25cm}
%   \caption{WER results on English ASR test sets from public datasets, including LibriSpeech (test-clean and test-other) and FLEURS, along with EER results on a series of in-house English ASR evaluation sets spanning diverse domains. domains. Note that the training data used for the models constructed in this paper do not include these public datasets.}
%   %\vspace{-0.05cm}
%   \label{tab:asr:main}
%   \centering
%   \setlength{\tabcolsep}{1.0mm}
%   \renewcommand\arraystretch{1.1}
%   \begin{tabular}{l | c c c c c c c c |c}
%     \Xhline{3\arrayrulewidth}
%     %\multirow{2}{*}{ST Model}&Quality&{Inference}&\multirow{2}{*}{Speedup}&Inference GPU&GPU Memory\\
%     %&(BLEU)&Time (s)&&Memory (MiB)&Reduction\\
%     % {ASR Model} & \multicolumn{3}{c|}{WER} & \multicolumn{8}{c}{EER} \\
%     % \cline{2-6}
%     {ASR Models}&Nutrition&Wellness&Clinical Meetings&Patient History&Pharmacy&Surgery&Banking&Medical&Avg\\
%     \hline
%     CoT-ASR&6.66&11.51&5.21&3.51&5.97&6.15&11.57&9.47&7.51\\
%     {\ } w/ user context&5.24&8.13&4.46&3.31&3.11&3.35&9.92&8.52&5.76\\
%     %NeurST \cite{DBLP:conf/acl/ZhaoWDYL21}&22.8&---&---&---&---\\
%     %Fairseq-ST \cite{DBLP:conf/ijcnlp/WangTMWOP20}&22.7&---&---&---&---\\
%     % MHA&23.18&281.3&1.00$\times$&18646&1.00\\
%     % MLA&22.97&97.0&2.90$\times$&5065&3.68\\
%     \Xhline{3\arrayrulewidth}
%   \end{tabular}
% %\vspace{-0.08cm}
% \end{table*}
\subsection{ASR Results with User-provided Context}
\begin{table}[ht!]
  \caption{
  EER ($\downarrow$) results on a series of in-house ASR evaluation sets, comparing CoT-ASR using its self-generated contextual reasoning (the default setting) with scenarios guided by user-provided context.}
  \label{tab:asr:user}
  \centering
  \setlength{\tabcolsep}{10pt}
  \renewcommand\arraystretch{1.05}
  \begin{tabular}{l|c|c}
    \Xhline{3\arrayrulewidth}
 \multirow{2}{*}{Test Set} & \multirow{2}{*}{CoT-ASR}&  CoT-ASR\\
    && with User Context\\
    \Xhline{2\arrayrulewidth}

    % \multirow{3}{*}{WER}
    %   & LibriSpeech test-clean & 2.41 & 2.20 \\
    %   & LibriSpeech test-other & 5.10 & 4.82 \\
    %   & FLEURS                & 3.85 &3.57\\

    \hline
     Consumer Goods  &13.46&9.03\\
      Nutrition           & 6.66& 5.24 \\
      Gaming        &14.53&12.99\\
      Wellness            & 11.51&8.13\\
      % Clinical Meetings    &  5.21 &4.46 \\
      Patient History       & 3.51 &3.31 \\
      Pharmacy             & 5.97 & 3.11\\
      Surgery             & 6.15 &3.35 \\
      Banking               & 11.57 &9.92 \\
      % Medical              & 9.47 &8.52 \\
      \cline{1-3}
      Average&9.17&6.89\\

    \Xhline{3\arrayrulewidth}
  \end{tabular}
\end{table}
This section compares the performance of CoT-ASR performance when generating transcriptions based on its self-inferred contextual analysis (the default setting) versus when guided by user-provided context, which allows CoT-ASR to skip the reasoning stage and leverage in-context learning to follow the user guide. As shown in Fig.~\ref{tab:asr:user}, 
the incorporation of user-guided context consistently yields further improvements over the already robust performance of CoT-ASR across all domain-specific test sets, resulting in an average relative reduction of 24.9\% in EER. This demonstrates that CoT-ASR possesses strong in-context learning capabilities and is able to effectively parse semantic cues from the context analysis to assist the subsequent transcription process.
The fact that variations in the preceding context analysis directly dictate the final transcription quality confirms that the model fundamentally benefits from the reasoning process, thereby validating that the reasoning-augmented ASR paradigm proposed in this paper represents a compelling and viable research direction.
Appendix~\ref{sec:error_analysis} provides a detailed analysis of CoT-ASR's robustness to errors in the context analysis phase.

\subsection{Ablation Studies On CTC-guided Modality Adapter}
\begin{table}[ht!]
  \caption{
  EER ($\downarrow$) results on a series of in-house ASR evaluation sets across multiple domains, comparing CoT-ASR with the proposed CTC-guided Modality Adapter (the default setting) against a linear-projection-based adapter (with activation function).}
  \label{tab:asr:ctc}
  \centering
  \setlength{\tabcolsep}{10pt}
  \renewcommand\arraystretch{1.05}
  \begin{tabular}{l|c|c}
    \Xhline{3\arrayrulewidth}
 \multirow{2}{*}{Test Set} & \multirow{2}{*}{CoT-ASR}&  CoT-ASR\\
    && with Linear Adapter\\
    \Xhline{2\arrayrulewidth}

    % \multirow{3}{*}{WER}
    %   & LibriSpeech test-clean & 2.41 & 2.20 \\
    %   & LibriSpeech test-other & 5.10 & 4.82 \\
    %   & FLEURS                & 3.85 &3.57\\

    \hline
      Consumer Goods  &13.46&14.93\\
      Nutrition           & 6.66& 6.80 \\
      Gaming        &14.53&16.52\\
      Wellness            & 11.51&12.87\\
      % Clinical Meetings    &  5.21 &5.03 \\
      Patient History       & 3.51 &4.11 \\
      Pharmacy             & 5.97 & 6.66 \\
      Surgery             & 6.15 &6.56 \\
      Banking               & 11.57 &12.28 \\
      % Medical              & 9.47 &10.34 \\
      \cline{1-3}
      Average&9.17&10.09\\

    \Xhline{3\arrayrulewidth}
  \end{tabular}
\end{table}
This section conducts ablation studies on the proposed CTC guided Modality Adapter and compares it with the commonly used linear projection-based adapter consisting of two linear layers and a GELU activation. Both adapters preserve the temporal length of the encoder outputs, which retains richer information to benefit ASR performance.

As shown in Table~\ref{tab:asr:ctc}, the CTC guided Modality Adapter consistently yields further improvements, achieving an average relative EER reduction of 9.1\% compared to the linear adapter. This gain can be attributed to the limitations of linear adapters, which rely solely on training time optimization to implicitly map encoder outputs from the speech modality into the LLM text latent space, but do not leverage the readily available LLM text embeddings, even though text based LLMs are pretrained directly on these embeddings.
In contrast, the proposed CTC-guided Modality Adapter explicitly exploits CTC posterior distributions to directly incorporate LLM embedding vectors, enabling a more efficient and effective speech to text modality adaptation. Appendix~\ref{sec:appendix_ctc_comparison} further compares it with other alignment-aware adapters.
%---
\subsection{Comparisons to Recent Advanced Models}

% \begin{table*}[ht!]
%   \caption{
%   EER ($\downarrow$) results on a series of in-house ASR evaluation sets, comparing CoT-ASR using its self-generated contextual reasoning (the default setting) with scenarios guided by user-provided context.
%   }
%   \label{tab:asr:recent}
%   \centering
%   \setlength{\tabcolsep}{5pt} % 适当增加间距使表格更大气
%   \renewcommand\arraystretch{1.2} % 稍微增加行高提高可读性
%   \begin{tabular}{l|c|c|c|c|c|c|c}
%     \Xhline{3\arrayrulewidth}
%     \textbf{LLM} & \multicolumn{2}{c|}{Phi-4} & {Whisper}& Gemma 3n&Qwen3 &Qwen2.5&Voxtral\\ 
%     \textbf{Size} & \multicolumn{2}{c|}{3.8B}&1.5B&8B&30B&7B&3B\\
%     \hline
%     \multirow{2}{*}{\textbf{Test Set}} & {Phi4MM-style} &\multirow{2}{*}{{CoT-ASR}} &{Whisper-}&{Gemma 3n }&Qwen3-Omni-&Qwen2.5-&Voxtral-Mini\\
%         &  {Baseline}   & & Large-v3& E4B IT&30B-A3B-Instruct&Omni-7B&-3B-2507 \\
%     \Xhline{2\arrayrulewidth}
%     Nutrition       &7.36  & 6.66  & 7.13 &11.77 \\
%     Wellness        &14.27  & 11.51&11.88 &26.15\\
%     Clinical Meet. &5.63 & 5.21&5.35 &9.56\\
%     Patient History   &4.31& 3.51  &4.36&6.94 \\
%     Pharmacy         &9.01 & 5.97 &6.80 &10.74 \\
%     Surgery          &7.44 & 6.15 &5.82 &8.54 \\
%     Banking          &12.99 & 11.57 &10.80 &20.78\\
%     Medical          &10.65 & 9.47 & 9.00 &18.73\\
%     \hline
%     \textbf{Average}  &8.96& 7.51&7.64 &14.15\\
%     \Xhline{3\arrayrulewidth}
%   \end{tabular}
% \end{table*}

\begin{table*}[ht!]
  \caption{
  EER ($\downarrow$) results on a series of in-house ASR evaluation sets across multiple domains, comparing CoT-ASR with several open-source foundation models, including Whisper-large-v3, Gemma 3n E4B IT, Qwen3-Omni-30B-A3B-Instruct, Qwen2.5-Omni-7B, and Voxtral-Mini-3B-2507. Note that the comparison is not well controlled and is provided for context, as differences in LLM backbones and training data prevent direct comparison. The models built in this paper were trained on 38k hours of data, whereas the open-source models are trained on substantially larger datasets, for example, Whisper-large-v3 used approximately 5M hours of training data. In all cases, speech serves as the sole input, without any additional context provided.
  }
  \label{tab:asr:recent}
  \centering
  \setlength{\tabcolsep}{0.7pt} 
  \renewcommand\arraystretch{1.1} 
  \begin{tabular}{l cc ccccc}
    \toprule
    \textbf{LLM} & \multicolumn{2}{c}{Phi4} & Whisper & Gemma 3n & Qwen3 & Qwen2.5 & Voxtral \\ 
    \textbf{Size} & \multicolumn{2}{c}{3.8B} & 1.5B & 8B & 30B & 7B & 3B \\
    \cmidrule(lr){2-3} \cmidrule(lr){4-4} \cmidrule(lr){5-5} \cmidrule(lr){6-6} \cmidrule(lr){7-7} \cmidrule(lr){8-8}
    
    \multirow{2}{*}{\textbf{Test Set}} & \makecell{Phi4MM-style\\Baseline} & {CoT-ASR} & \makecell{Whisper-\\Large-v3} & \makecell{Gemma 3n\\E4B IT} & \makecell{Qwen3-Omni-\\30B-A3B-Instruct} & \makecell{Qwen2.5-\\Omni-7B} & \makecell{Voxtral-Mini\\-3B-2507} \\
    \midrule
    Consumer.   &15.45  &13.46  &14.32  &22.66 & 15.02 &23.79&14.32\\
    Nutrition        & 7.36  & 6.66  & 7.13 & 11.77 & 6.63 & 14.72 & 7.36 \\
    Gaming           & 17.42 & 14.53 & 15.14 & 23.59 & 15.01 & 28.68 & 14.15 \\
    Wellness         & 14.27 & 11.51 & 11.88 & 26.15 & 11.51 & 23.76 & 13.31 \\
    % Clinical Meet.   & 5.63  & 5.21  & 5.35 & 9.56  & 4.49 & 6.74 & 4.78 \\
    Patient.  & 4.31  & 3.51  & 4.36 & 6.94  & 4.05 & 8.14 & 4.56\\
    Pharmacy         & 9.01  & 5.97  & 6.80 & 10.74 & 3.45 & 9.18 & 7.39\\
    Surgery          & 7.44  & 6.15  & 5.82 & 8.54  & 5.76 & 14.79 & 6.87\\
    Banking          & 12.99 & 11.57 & 10.80 & 20.78 & 12.06& 21.26 & 12.45 \\
    % Medical          & 10.65 & 9.47  & 9.00 & 18.73 & -- & 17.18 & 9.30 \\
    \midrule
    \textbf{Average} & 11.03  & 9.17 & 9.53 & 16.40 & 9.19 & 18.04 & 10.05 \\
    \bottomrule
  \end{tabular}
\end{table*}

This section experimentally compares the proposed CoT-ASR with a set of strong open-source foundation models, including Whisper-large-v3 \cite{pmlr-v202-radford23a}, Gemma 3n E4B IT \cite{team2025gemma}, Qwen3-Omni-30B-A3B-Instruct \cite{Xu2025Qwen3OmniTR}, Qwen2.5-Omni-7B \cite{xu2025qwen2}, and Voxtral-Mini-3B-2507 \cite{liu2025voxtral}. These models are not directly comparable to ours due to differences in LLM backbones and training data, and thus the comparison is not well controlled and serve as a context for our results. The models built in this paper were trained on 38k hour speech data, which is substantially less than the training data used by these open-source models. 
%As shown in Table~\ref{tab:asr:recent}, models with much larger parameter sizes, such as Qwen3-Omni-30B-A3B-Instruct, Gemma 3n E4B IT, and Qwen2.5-Omni-7B, do not exhibit a proportional performance advantage over smaller models. This indicates that, for ASR tasks, the strong capabilities of LLMs are not easily translated into ASR gains, which directly motivates the reasoning-based ASR paradigm proposed in this paper.
As shown in Table~\ref{tab:asr:recent}, much larger models, such as Qwen3-Omni-30B-A3B-Instruct, Gemma 3n E4B IT, and Qwen2.5-Omni-7B, do not show a proportional performance gain over smaller models. This suggests that for ASR, the strong abilities of LLMs do not easily translate into improvements, motivating the reasoning-based ASR paradigm proposed in this paper.

%From the results, Whisper-large-v3, as a 1.5B-parameter model, demonstrates relatively robust performance; however, it is the only model based on an AED architecture rather than an LLM-based ASR framework. Voxtral-Mini-3B-2507, a 3B-parameter model, achieves better performance than Gemma 3n E4B IT and Qwen2.5-Omni-7B. Qwen3-Omni-30B-A3B-Instruct, as the largest model, shows strong ASR performance and outperforms other open-source models, achieving an average relative EER reduction of 3.6\% compared to Whisper-large-v3.

Whisper-large-v3, with 1.5B parameters, shows relatively robust performance but is the only AED-based model rather than LLM-based. Voxtral-Mini-3B-2507 outperforms Gemma 3n E4B IT and Qwen2.5-Omni-7B. The largest model, Qwen3-Omni-30B-A3B-Instruct, achieves the strongest ASR results, with an average relative EER reduction of 3.6\% over Whisper-large-v3.

For the models built in this paper, the Phi4MM-style Baseline model underperforms strong open-source ASR models such as Whisper-large-v3, whereas CoT-ASR surpasses these models and even achieves a slightly lower average EER than the strongest open-source model, Qwen3-Omni-30B-A3B-Instruct. 
%These results demonstrate the superior ASR performance of CoT-ASR.
These results demonstrate the superior efficiency and effectiveness of the CoT-ASR framework.

\subsection{Comparisons to Contextual Biasing}
\label{sec:biasing_librispeech}
This section provides an additional comparison between CoT-ASR and contextual biasing. 
Our Phi4MM-style baseline was further prompted to use the provided biasing list with: “Transcribe the audio clip into text with extra attention to the following words: [biasing list]” and was compared with our proposed CoT-ASR.
Note that this comparison is inherently unfair, as CoT-ASR does not use biasing lists as input and relies solely on speech input, while biasing lists are not always available in practice. We include this comparison to provide additional insights.
For reproducibility, this section uses the publicly available biasing lists from the fbai-speech \cite{DBLP:conf/interspeech/LeJKKSMCSFKSS21} project\footnote{\url{https://github.com/facebookresearch/fbai-speech}} on LibriSpeech, along with its official evaluation metrics, including WER, B-WER (biased WER), and U-WER (unbiased WER).

The results are shown in Table~\ref{tab:librispeech_biasing_vertical}. Without biasing, CoT-ASR consistently outperforms the baseline across both test sets. When the biasing list of size 100 is used, the baseline achieves lower B-WER, indicating improved recognition of bias-targeted words. However, this comes at the cost of higher U-WER, reflecting worse performance on general vocabulary. As a result, the overall WER does not improve and can even degrade significantly with a larger bias list (e.g., size 500).
These results suggest a potential trade-off in biasing-based approaches: improving recognition of targeted words can sometimes come at the expense of general recognition. In contrast, CoT-ASR achieves improvements across both frequent and rare words without relying on external biasing inputs, offering a more balanced performance.
% These results highlight a key trade-off in biasing-based approaches: improving recognition of targeted words may harm general recognition performance. In contrast, CoT-ASR improves recognition holistically without relying on external biasing inputs, leading to more balanced performance across both frequent and rare words.

\begin{table}[t]
\centering
\caption{WER / B-WER (biased WER) / U-WER (unbiased WER) ($\downarrow$) results on LibriSpeech test sets, comparing CoT-ASR with Phi4MM-style baseline using contextual biasing. Note that this comparison is not well-controlled, as CoT-ASR does not receive the biasing list as additional input.}
\label{tab:librispeech_biasing_vertical}
\begin{tabular}{lcccccc}
\toprule
\textbf{Model} & \multicolumn{3}{c}{\textbf{test-clean}} & \multicolumn{3}{c}{\textbf{test-other}} \\
\cmidrule(lr){2-4} \cmidrule(lr){5-7}
 & WER & B-WER & U-WER & WER & B-WER & U-WER \\
\midrule
CoT-ASR & 2.20 & 9.31 & 1.29 & 4.82 & 20.01 & 3.02 \\
%\hline
Phi4MM-style Baseline & 2.41 & 10.41 & 1.40 & 5.10 & 21.33 & 3.18 \\
%\addlinespace
{\ \ \ }+Bias List (Size=100) & 2.42 & 8.84 & 1.60 & 5.56 & 18.47 & 4.03 \\
{\ \ \ }+Bias List (Size=500) & 5.99 & 10.83 & 5.25 & 13.80 & 22.51 & 12.78 \\
\bottomrule
\end{tabular}
\end{table}

\section{Conclusions}
\label{conclusion}
This paper introduces CoT-ASR, the first ASR model to integrate chain-of-thought (CoT) reasoning, establishing a reasoning-based ASR paradigm that leverages LLMs’ rich knowledge and contextual understanding to generate more informed transcriptions.
%This paper proposes CoT-ASR, the first work to integrate CoT reasoning into ASR to enable a reasoning-based ASR model. CoT-ASR represents a new ASR paradigm that more effectively exploits the rich knowledge and strong contextual understanding of LLMs to produce more informed transcriptions. 
By explicitly constructing a reasoning chain, CoT-ASR performs contextual reasoning and transcription in a single one-pass generation, maintaining the simplicity and general design of speech LLM pipelines. Furthermore, CoT-ASR naturally supports user-guided transcription: 
user-provided context can be incorporated at inference time to bypass self-generated reasoning,
% user-provided context can bypass self-generated reasoning, 
allowing in-context learning to produce higher-quality transcriptions.
% Furthermore, CoT-ASR naturally supports user-guided transcription: user-provided context can be incorporated at inference time to bypass self-generated reasoning and leverage in-context learning, leading to higher-quality transcriptions under explicit guidance. 
This paper also introduces a CTC-guided Modality Adapter, which efficiently adapts speech to the LLM latent space by weighting LLM embeddings according to CTC non-blank probabilities.
%which facilitates efficient modality adaptation from speech to the LLM's latent space by using CTC non-blank probabilities to guide a weighted aggregation of corresponding LLM embedding vectors. 
Experiments show that CoT-ASR improves ASR results over standard LLM-based ASR approaches.

\iffalse
\begin{ack}
Use unnumbered first level headings for the acknowledgments. All acknowledgments
go at the end of the paper before the list of references. Moreover, you are required to declare
funding (financial activities supporting the submitted work) and competing interests (related financial activities outside the submitted work).
More information about this disclosure can be found at: \url{https://neurips.cc/Conferences/2026/PaperInformation/FundingDisclosure}.

Do {\bf not} include this section in the anonymized submission, only in the final paper. You can use the \texttt{ack} environment provided in the style file to automatically hide this section in the anonymized submission.
\end{ack}
\fi

\bibliographystyle{elsarticle-harv}
\bibliography{ref}

\iffalse
\section*{References}

References follow the acknowledgments in the camera-ready paper. Use unnumbered first-level heading for
the references. Any choice of citation style is acceptable as long as you are
consistent. It is permissible to reduce the font size to \verb+small+ (9 point)
when listing the references.
Note that the Reference section does not count towards the page limit.
\medskip

{
\small

[1] Alexander, J.A.\ \& Mozer, M.C.\ (1995) Template-based algorithms for
connectionist rule extraction. In G.\ Tesauro, D.S.\ Touretzky and T.K.\ Leen
(eds.), {\it Advances in Neural Information Processing Systems 7},
pp.\ 609--616. Cambridge, MA: MIT Press.

[2] Bower, J.M.\ \& Beeman, D.\ (1995) {\it The Book of GENESIS: Exploring
  Realistic Neural Models with the GEneral NEural SImulation System.}  New York:
TELOS/Springer--Verlag.

[3] Hasselmo, M.E., Schnell, E.\ \& Barkai, E.\ (1995) Dynamics of learning and
recall at excitatory recurrent synapses and cholinergic modulation in rat
hippocampal region CA3. {\it Journal of Neuroscience} {\bf 15}(7):5249-5262.
}

\fi
%%%%%%%%%%%%%%%%%%%%%%%%%%%%%%%%%%%%%%%%%%%%%%%%%%%%%%%%%%%%
%\newpage
\appendix
\section{Prompt Templates}
\label{app:prompt}
Table~\ref{tab:prompts} shows the prompt templates used to generate the reasoning-augmented ASR training data.
\begin{table}[H]
    \centering
    \caption{Prompt used in this paper to construct reasoning-augmented ASR data.}
    % \vspace{0.1in}
    \small
    \renewcommand\arraystretch{0.98} 
    \begin{tabular}{lp{13cm}}
    \toprule
    & Prompt Content \\
    \midrule
    & $<$INSTRUCTION$>$ \\
    & You are a High-Fidelity Context Synthesizer. Your task is to generate a $<$CONTEXT$>$ block based on the provided $<$GROUND\_TRUTH$>$ text. \\
    \\
    & This context will be placed before the ASR transcript to guide a Speech-LLM. It must act as a semantic "warm-up" by defining the environment, the speaker's intent, and the specific entities or examples mentioned. \\
    \\
    & CORE ATTRIBUTES OF THE CONTEXT: \\
    & 1. Scenario Identification: Start by identifying the format (e.g., educational lecture, formal briefing, casual discussion). \\
    & 2. Specific Weaving: Incorporate unique examples, technical terms, or key names found in the text (e.g., if the text mentions "sugar" or "Sherman", the context must include them) to ground the model. \\
    & 3. Relational Dynamics: Describe who is speaking and to whom (e.g., "addressing a group", "referencing an absent colleague"). \\
    & 4. No Spoilers/Direct Quotes: Describe what is being discussed rather than just repeating the transcript verbatim. \\
    \\
    & OUTPUT FORMAT: $<$CONTEXT$>$[Your 2-3 sentence analysis]$<$/CONTEXT$>$ $<$TRANSCRIPT$>$[The exact ground truth text]$<$/TRANSCRIPT$>$ \\
    \\
    & $<$FEW\_SHOT\_EXAMPLES$>$ \\
    & \\
    & Example 1: Trading/Education \\
    & $<$INPUT$>$ \\
    & $<$GROUND\_TRUTH$>$So this is what would be known as the speculators...$<$/GROUND\_TRUTH$>$ \\
    &$<$/INPUT$>$ \\
    & $<$OUTPUT$>$ \\
    & $<$CONTEXT$>$This is an educational context where the speaker is explaining the differences between physical and cash settlements...$<$/CONTEXT$>$$<$TRANSCRIPT$>$So this is what would be known as the speculators...$<$/TRANSCRIPT$>$ \\
    & $<$/OUTPUT$>$ \\
    & \\
    & Example 2: Policy/Discussion \\
    & $<$INPUT$>$ \\
    & $<$GROUND\_TRUTH$>$For instance, if you got a tax cut at the federal level, but because of record budget shortfalls in state and local government budgets...$<$/GROUND\_TRUTH$>$ \\
    & $<$/INPUT$>$ \\
    & $<$OUTPUT$>$ \\
    & $<$CONTEXT$>$This is a discussion on fiscal policy and public opinion...$<$/CONTEXT$>$ $<$TRANSCRIPT$>$For instance, if you got a tax cut at the federal level...$<$/TRANSCRIPT$>$ \\
    & $<$/OUTPUT$>$ \\
    & \\
    & Example 3: Diplomacy/Press \\
    & $<$INPUT$>$ \\
    & $<$GROUND\_TRUTH$>$So as you know, Under Secretary Sherman was there; she's had conversations with these leaders...$<$/GROUND\_TRUTH$>$ \\
    & $<$/INPUT$>$ \\
    & $<$OUTPUT$>$ \\
    & $<$CONTEXT$>$This is a formal press conference or briefing context involving high-level political discussions...$<$/CONTEXT$>$ $<$TRANSCRIPT$>$So as you know, Under Secretary Sherman was there...$<$/TRANSCRIPT$>$ \\
    & $<$/OUTPUT$>$ \\
    &\\
    & $<$/FEW\_SHOT\_EXAMPLES$>$ \\
    &\\
    & $<$THE\_TASK$>$ \\
    &Generate the $<$CONTEXT$>$ and $<$TRANSCRIPT$>$ for the following $<$GROUND\_TRUTH$>$ text: \\
    & $<$INPUT$>$ \\
    & $<$GROUND\_TRUTH$>${text}$<$/GROUND\_TRUTH$>$ \\
    & $<$/INPUT$>$ \\
    & $<$/THE\_TASK$>$ \\
    \bottomrule
    \end{tabular}
    \label{tab:prompts}
\end{table}

\section{Inference Efficiency}
\label{sec:efficiency}
CoT-ASR, as a reasoning-based ASR model, generates additional tokens for the chain-of-thought reasoning, which introduces some computational overhead. Since our models are designed for offline ASR, this additional latency is not critical. Modern GPUs and techniques such as parallel decoding can further mitigate this overhead. 
In our experiments on the Consumer Goods test set (H200 GPU), the Phi4MM-style baseline has a real-time factor (RTF) of 0.157, whereas CoT-ASR with the CTC-guided adapter has an RTF of 0.257. For comparison, CoT-ASR using the simpler linear adapter shows a similar RTF of 0.253, indicating that the choice of adapter has minimal impact on decoding speed.
These measurements are based on PyTorch decoding; switching to a framework such as vLLM is expected to further reduce RTF. 
Note that the term `efficient'' used in Section~\ref{sec:ctc-adapter} regarding the CTC-guided Modality Adapter refers to the efficiency of modality adaptation rather than computational speed.

\section{Expanded Results: Main Experiments}
\label{sec:appendix_internal_wer}

In addition to the main evaluation using Entity Error Rate (EER), this section also reports Word Error Rate (WER) on our in-house test sets to provide a broader perspective on the general transcription quality. While these in-house sets are primarily designed for entity-level evaluation, analyzing WER can assess overall recognition performance beyond critical entities.
Table~\ref{tab:internal_wer} summarizes the WER results of our proposed CoT-ASR compared with the strong Phi4MM-style baseline. CoT-ASR consistently reduces WER across all domains compared to the baseline. This demonstrates that, in addition to its entity-level advantages, CoT-ASR also improves general transcription quality, reflecting its design as a robust and general-purpose ASR system. These results further support the paradigm shift introduced by CoT-ASR, where reasoning-based processing enables both accurate entity recognition and overall transcription improvements.

\begin{table}[h]
\centering
\caption{WER ($\downarrow$) results on a series of in-house English ASR evaluations sets.}
\label{tab:internal_wer}
\begin{tabular}{lcc}
\toprule
\textbf{Test Set} & \textbf{Phi4MM-style Baseline} & \textbf{CoT-ASR} \\
\midrule
Consumer Goods & 7.38 & 6.47 \\
Nutrition & 5.00 & 4.67 \\
Gaming & 8.56 & 8.03 \\
Wellness & 9.13 & 8.31 \\
Patient History & 2.42 & 2.08 \\
Pharmacy & 2.54 & 2.22 \\
Surgery & 2.45 & 2.29 \\
Banking & 9.37 & 8.18 \\
\midrule
\textbf{Average} & \textbf{5.86} & \textbf{5.28} \\
\bottomrule
\end{tabular}
\end{table}

%\appendix
% \section{Additional Analysis: Inference Efficiency and User Context Robustness}
% \label{sec:appendix_efficiency_robustness}

\section{Robustness to Contextual Analysis Errors}
\label{sec:error_analysis}

To evaluate the robustness of CoT-ASR to errors during the contextual analysis stage, this section introduces controlled errors into the user-provided context by randomly inserting, deleting, or substituting tokens at varying probabilities. This controlled setup allows for a systematic assessment, since CoT-ASR's default self-reasoning mode generates context internally, which is less straightforward to manipulate. Table~\ref{tab:user_context_robustness} reports the resulting EER on our in-house ASR evaluation sets. The results show that while errors in the contextual analysis phase can affect transcription, CoT-ASR maintains reasonably strong performance even with 50\% simulated noise. Note that in its default setting, CoT-ASR relies solely on self-generated reasoning from the speech input and does not require external user-provided context.

\begin{table}[h]
\centering
\caption{Robustness of CoT-ASR to simulated errors in user-provided context. EER (\%) is reported under varying error rates.}
\label{tab:user_context_robustness}
\begin{tabular}{lcccc}
\toprule
\textbf{Test Set} & \textbf{0\% error} & \textbf{10\% error} & \textbf{25\% error} & \textbf{50\% error} \\
\midrule
Consumer Goods & 9.03 & 11.02 & 11.46 & 12.76 \\
Nutrition & 5.24 & 5.44 & 5.67 & 6.07 \\
Gaming & 12.99 & 13.70 & 14.50 & 14.24 \\
Wellness & 8.13 & 8.50 & 9.70 & 10.65 \\
Patient History & 3.31 & 3.54 & 3.47 & 3.58 \\
Pharmacy & 3.11 & 3.31 & 3.94 & 4.38 \\
Surgery & 3.35 & 4.32 & 4.81 & 4.66 \\
Banking & 9.92 & 10.53 & 10.86 & 11.51 \\
\midrule
\textbf{Average} & \textbf{6.89} & \textbf{7.55} & \textbf{8.05} & \textbf{8.48} \\
\bottomrule
\end{tabular}
\end{table}

\section{Comparison with CTC Compressor and other Alignment-aware Adapter}
\label{sec:appendix_ctc_comparison}

This section provides additional comparisons between the proposed CTC-guided Modality Adapter and existing alignment-ware adapters, including the CTC compressor \cite{fan2025alignformer} and SSR \cite{tan-etal-2025-ssr}.
Prior work \cite{fathullah2024prompting} as well as our large-scale industrial pre-experiments show that, for LLM-based ASR, even slight compression of the speech prompt length can lead to noticeable performance degradation. This motivates the design of our Phi4MM-style baseline, which adopts a linear adapter without any down-sampling and already serves as a strong baseline.

Existing approaches such as CTC compressors \cite{fan2025alignformer} and alignment-aware speech-text connectors \cite{tan-etal-2025-ssr} typically rely on speech-text alignment to compress the speech prompt so that its length better matches the text sequence. While intuitive, such compression inevitably discards information and harms ASR performance. In contrast, the proposed CTC-guided Modality Adapter does not compress the speech prompt. Instead, it directly maps speech features into the LLM text embedding space by leveraging existing LLM embedding vectors and weighting them according to CTC non-blank probabilities. Unlike prior methods that learn this mapping gradually during training, our approach performs a more direct transformation. Moreover, by explicitly distinguishing blank and non-blank probabilities (Eqs.~\ref{eq:b}-\ref{eq:b_hat}), even blank-dominant frames are preserved, retaining more information.

\begin{table}[H]
\centering
\caption{EER ($\downarrow$) results on on in-house ASR evaluation sets for CoT-ASR with different modality adapters.}
\label{tab:ctc_comparison}
\begin{tabular}{lcccc}
\toprule
\textbf{Test Set} & \textbf{CoT-ASR} & \textbf{w/ Linear Adapter} & \textbf{w/ CTC Compressor} & \textbf{w/ SSR} \\
\midrule
Consumer Goods & 13.46 & 14.93 & 21.70 & 21.44 \\
Nutrition & 6.66 & 6.80 & 11.57 & 12.23 \\
Gaming & 14.53 & 16.52 & 24.29 & 25.12 \\
Wellness & 11.51 & 12.87 & 19.12 & 18.37 \\
Patient History & 3.51 & 4.11 & 7.03 & 7.16 \\
Pharmacy & 5.97 & 6.66 & 11.57 & 10.88 \\
Surgery & 6.15 & 6.56 & 12.76 & 13.30 \\
Banking & 11.57 & 12.28 & 17.32 & 17.38 \\
\midrule
\textbf{Average} & \textbf{9.17} & \textbf{10.09} & \textbf{15.67} & \textbf{15.73} \\
\bottomrule
\end{tabular}
\end{table}

The results in Table~\ref{tab:ctc_comparison} show that the linear adapter is already a strong baseline, as it preserves the full speech prompt without compression. In contrast, both the CTC compressor \cite{fan2025alignformer} and SSR \cite{tan-etal-2025-ssr} degrade performance due to information loss caused by prompt compression. The proposed CTC-guided Modality Adapter retains the full speech prompt and performs efficient modality mapping based on CTC non-blank probabilities, leading to consistent performance gains. 

Overall, although these methods all involve CTC, the proposed CTC-guided Modality Adapter follows a fundamentally different design philosophy. Rather than compressing the speech prompt to match text length, it preserves the full prompt and focuses on effective mapping into the LLM embedding space.

%%%%%%%%%%%%%%%%%%%%%%%%%%%%%%%%%%%%%%%%%%%%%%%%%%%%%%%%%%%%

\newpage

\end{document}